# Extended monocular 3D imaging

Zicheng Shen[1,#], Feng Zhao[1#,*], Yibo Ni[1], Yuanmu Yang[1*]

[1]State Key Laboratory of Precision Measurement Technology and Instruments, Department of Precision Instrument, Tsinghua University, Beijing 100084, China

\# These authors contributed equally.

\* Email: fengzhao@mail.tsinghua.edu.cn; ymyang@tsinghua.edu.cn

**3D vision is of paramount importance for numerous applications ranging from machine intelligence to precision metrology. Despite much recent progress, the majority of 3D imaging hardware remains bulky and complicated and provides much lower image resolution compared to their 2D counterparts. Moreover, there are many well-known scenarios that existing 3D imaging solutions frequently fail. Here, we introduce an extended monocular 3D imaging (EM3D) framework that fully exploits the vectorial wave nature of light. Via the multi-stage fusion of diffraction- and polarization-based depth cues, using a compact monocular camera equipped with a diffractive-refractive hybrid lens, we experimentally demonstrate the snapshot acquisition of a million-pixel and accurate 3D point cloud for extended scenes that are traditionally challenging, including those with low texture, being highly reflective, or nearly transparent, without a data prior. Furthermore, we discover that the combination of depth and polarization information can unlock unique new opportunities in material identification, which may further expand machine intelligence for applications like target recognition and face anti-spoofing. The straightforward yet powerful architecture thus opens up a new path for a higher-dimensional machine vision in a minimal form factor, facilitating the deployment of monocular cameras for applications in much more diverse scenarios.**

Three-dimensional (3D) vision is essential for machines and artificial intelligence to perceive and interact with the world. As application scenarios widen, existing 3D imaging solutions, primarily including time-of-flight, structured light, and multi-view stereo, face numerous challenges[1-5]. For instance, time-of-flight[6] and structured light[7]-based 3D sensors require active laser illumination and often suffer from limited image resolution. In addition, they generally perform poorly in measuring highly reflective and nearly transparent objects. Binocular or multi-view cameras[8] operate without active laser illumination, yet they also have a well-known limitation in measuring targets with low texture. Furthermore, their depth estimation accuracy is constrained by the baseline length, resulting in a larger form factor and a stringent calibration requirement.

Monocular depth estimation has been a major research focus in the computer vision community for over a decade. In that context, the hardware can be as simple as a 2D camera, with 3D information inferred from the 2D image via learning[9,10]. However, most deep learning models can only provide relative depth trends with limited accuracy. Their scalability and real-world applicability are constrained by the available training datasets. Alternatively, one can leverage the concept of depth-from-defocus (DfD)[11-14] of a monocular camera to derive absolute depth information from axial image blur. By designing more sophisticated depth-dependent point-spread functions (PSFs), which can be generated using diffractive optical elements (DOEs) or metasurfaces, the depth estimation accuracy can be further improved[15-23]. However, similar to binocular cameras, the DfD method also encounters challenges when measuring low-texture targets. With limited input information, there are often voids and errors in the calculated 3D image, making detailed 3D reconstruction of complex surfaces extremely difficult.

While most existing 3D imaging methods only treat light as a scalar field, a new 3D imaging modality, namely shape-from-polarization (SfP)[24-26], has recently emerged, which can further leverage the vectorial property of light waves to enable the detailed reconstruction of 3D surfaces even for low-texture objects. The SfP method estimates surface normal by analyzing both the polarization angle and degree-of-polarization of light reflected from the target object. However, a major limitation of SfP is that it can only provide relative depth information. More importantly, SfP is inherently an ill-posed problem: there is an ambiguity in the normal vector determination owing to the multi-valued nature of trigonometric functions. Therefore, to mitigate the ambiguity and achieve reliable 3D reconstruction, SfP needs to work in conjunction with other 3D imaging techniques including the aforementioned time-of-flight[27], structured light[28], or multi-view stereo[29], making the overall system even more bulky and complicated, limiting its broader deployment.

Here, we propose an extended monocular 3D imaging (EM3D) apparatus that allows high-quality, snapshot 3D imaging across extended challenging scenes by fully exploiting the vectorial wave nature of light. Through the multi-stage fusion of diffraction- and polarization-based depth cues, we experimentally demonstrate the reconstruction of detailed 3D surfaces with precise absolute depth for a variety of target objects, including those with low texture, high complexity, or being highly reflective or nearly transparent. Furthermore, we discover that the synergy of depth and polarization information can unlock unique new opportunities unattainable with depth or polarization alone, such as material identification, towards applications in target recognition and face anti-spoofing.

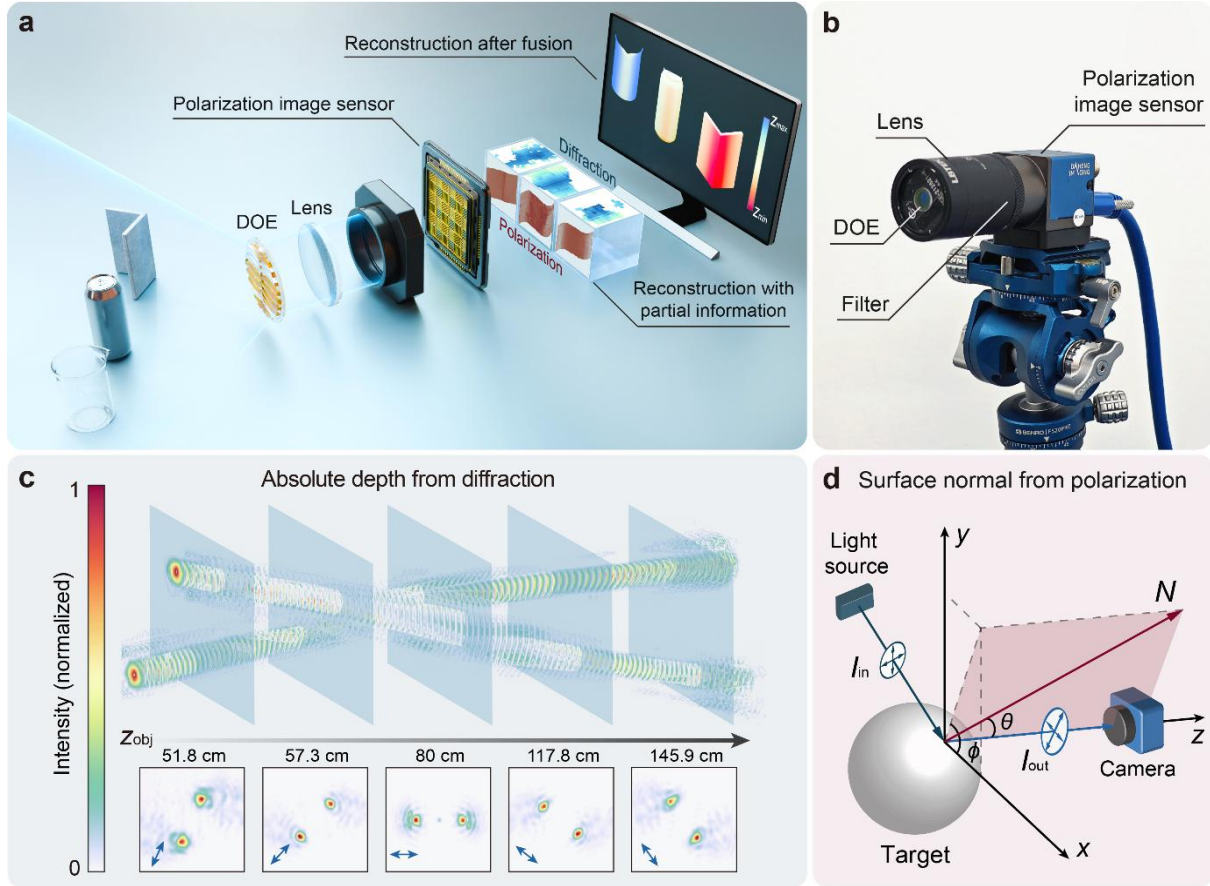

**Figure 1 | Framework of the extended monocular 3D imaging system. a,** Framework of the extended monocular 3D imaging system that works for a wide range of challenging scenes, including a low-texture corner, a reflective metal jar, and a nearly transparent glass beaker. The monocular camera is equipped with a diffractive-refractive hybrid lens to provide an absolute yet rough depth map and a polarization image sensor to provide detailed yet ambiguous surface normal determination. The fusion of both depth cues results in a detailed and accurate 3D surface reconstruction with absolute depth information. **b,** Photograph of the camera prototype. **c,** Depth cue (absolute depth) from diffraction. Simulated (top panel) and experimentally measured (bottom panel) depth-dependent double-helix PSF of the diffractive-refractive hybrid lens. **d,** Depth cue (surface normal) from polarization. $I_{in}$ and $I_{out}$ are incident and reflected light, respectively. $x$, $y$, and $z$ are spatial coordinates where the $z$-axis is along the direction of $I_{out}$, and $x$-$y$ plane is parallel to the image plane. $N$ is the surface normal of the object, with $\theta$ and $\phi$ being the zenith and azimuth angle, respectively. $N$ can be derived from the polarization state of the reflected light.

**Framework of the extended monocular 3D imaging system**

The framework of the monocular 3D imaging system is schematically illustrated in Fig. 1a. The camera hardware is composed of a diffractive-refractive hybrid lens, a bandpass filter, and a polarization image sensor (Sony IMX250). The assembled prototype has a size of $3.1 \times 3.6 \times 8$ cm$^3$ (Fig. 1b). The DOE, which is fabricated by three-step photo-lithography and is readily mass-producible via nanoimprinting[30,31] (Methods and Supplementary Section 1), is designed to generate a depth-dependent double-helix PSF[32,33] (Supplementary Section 2). The double-helix PSF features two foci rotating around a central point, with the rotation angle depending on the axial depth of the object point. The simulated and experimentally measured PSFs of the diffractive-refractive hybrid lens (Methods and Supplementary Section 3) are shown in Fig. 1c. Given that the image is formed through the convolution between the object and the PSF, the depth of the object can be determined by the rotation angle of the double-helix PSF retrieved through cepstrum analysis[18,34]. Although the double-helix PSF has been widely used for 3D imaging in both micro- and macroscopic scenes[16,18,19,35], a well-known bottleneck is that it performs poorly when there is little texture in the target object, resulting in voids in the reconstructed 3D surface.

On the other hand, the SfP technique can measure the surface normal and reconstructs the 3D surface even for low-texture target objects. The surface normal of an object $N$ can be described by,

$$N = [\tan\theta \cos\phi, \tan\theta \sin\phi, 1], \tag{1}$$

where $\theta$ and $\phi$ are the zenith and azimuth angle, respectively. Based on the diffuse reflection model (Supplementary Section 4), as shown in Fig. 1d, $\theta$ can be inferred from the degree-of-linear-polarization ($DOLP$) of the light reflected from the surface as,

$$DOLP = \frac{\left(n_s - \frac{1}{n_s}\right)^2 \sin^2\theta}{2 + 2n_s^2 - \left(n_s + \frac{1}{n_s}\right)^2 \sin^2\theta + 4\cos\theta\sqrt{n_s^2 - \sin^2\theta}}, \tag{2}$$

where we define $n_s$ as the surface index of the target object, a parameter inherited from but differs from the refractive index $n$ in Fresnel equations. $n_s$ is often an unknown fitting parameter and may lead to inaccuracy in the zenith angle estimation[28]. Using the polarization image sensor equipped in the monocular 3D imaging system, $DOLP$ can be determined as,

$$DOLP = \frac{\sqrt{(I_{90} - I_0)^2 + (I_{45} - I_{135})^2}}{I_0 + I_{90}}, \tag{3}$$

where $I_0$, $I_{45}$, $I_{90}$, and $I_{135}$ are intensity measurements at polarization angles of 0°, 45°, 90° and 135°, respectively.

Meanwhile, $\phi$ can be calculated as,

$$\phi = \frac{1}{2}\arctan\left(\frac{I_0 + I_{90} - 2I_{45}}{I_{90} - I_0}\right) + 90°, \tag{4}$$

which is related to the polarization angle of the light reflected from the surface $\phi_0$ as $\phi = \phi_0 \pm 90°$. Due to the inherent ambiguity of $\pi$ radians of the arctan function, $\phi$ cannot be calculated deterministically[28]. Therefore, to achieve unambiguous polarization-based 3D reconstruction, it is necessary to use additional information to resolve the error in the surface normal (zenith and azimuth angle) determination.

Despite that both diffraction- and polarization-based depth cues have limitations, we find they happen to be complementary to each other and, more importantly, can be simultaneously obtained using a compact monocular camera. By developing a multi-stage fusion algorithm to facilitate the synergy between the two depth cues, the EM3D framework allows for detailed and accurate 3D imaging across a wide range of challenging scenes.

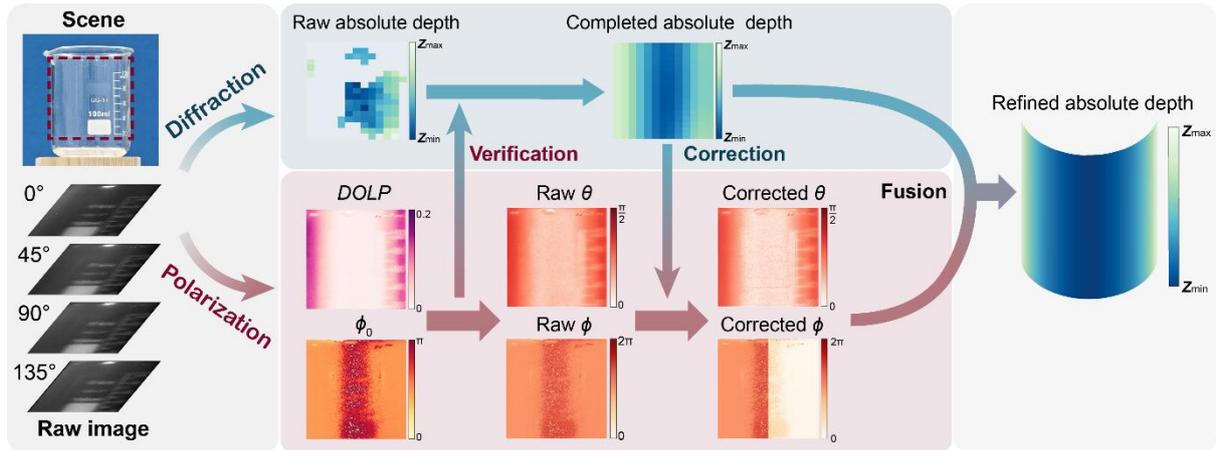

**Figure 2 | Multi-stage depth cue fusion algorithm.** Using a nearly transparent glass beaker as an example challenging scene, with the reconstructed region boxed by a red dashed line, the proposed monocular camera captures four polarization images in a snapshot. Initially, a rough absolute depth map with low confidence is computed using the limited and weak textures of the glass beaker based on the diffraction-based depth cue. The raw absolute depth map is refined by verifying depth trend continuity through polarization analysis, though it remains to have a low resolution. Since 3D reconstruction from polarization alone is ambiguous and can lead to a distorted surface shape, we use the completed absolute depth map to correct the zenith and azimuth angle derived from the polarization data, resulting in an accurate and detailed surface normal distribution of the glass beaker. Combining this refined surface normal distribution with the low-resolution depth map, a detailed and accurate absolute depth map of the glass beaker can be obtained.

The 3D image reconstruction flow is illustrated in Fig. 2. We use a nearly transparent glass beaker as an example scene, which is a typical challenging scene for most existing 3D imaging methods. The monocular camera captures an image that can be decomposed into four polarization images along 0°, 45°, 90°, and 135°, respectively, all of which are convoluted with the depth-dependent PSF. The depth-dependent PSF can be first used to generate a rough absolute depth map via analyzing the power cepstrum of sub-images[18,34]. However, due to the weak texture of the glass beaker, a significant portion of the calculated depth map has a low confidence level that is typically treated as void regions. To address the issue, the distribution of $DOLP$ and $\phi_0$ are calculated from the four polarization images and used to verify the depth trend continuity of the target object using a threshold that evaluates the continuity of the $DOLP$ distribution. Similar to the cost aggregation algorithm[36] used in stereo vision, we can fill in the void regions of the rough absolute depth map within the continuous surface through interpolation, despite that the resolution is still limited.

In the next stage, we leverage the completed absolute depth map to resolve the error and ambiguity of the surface normal derived from the polarization information. The surface azimuth and zenith of the completed depth map are first calculated from the depth gradient. Next, we determine whether to add a π phase to the azimuth angle $\phi$ derived from the polarization information by minimizing the difference between the azimuth angle $\phi$ calculated from polarization data and the completed absolute depth map[28]. Meanwhile, the surface index $n_s$ of the target object can be determined by minimizing the sum of absolute errors of zenith angle $\theta$ calculated from the two depth cues. Finally, the refined 3D surface shape can be combined with the low-resolution absolute depth map to generate a detailed and accurate depth map of the glass beaker with up to a million pixels (Supplementary Section 5).

**3D imaging in extended challenging scenes**
The 3D imaging performance of the proposed monocular camera system is experimentally tested across a wide variety of challenging scenarios, which include a cardboard box with minimal texture (Fig. 3a), a highly reflective metal jar (Fig. 3b), a nearly transparent glass beaker (Fig. 3c), and a human face with complex shape (Fig. 3d). A scene with minimal texture is challenging for conventional passive 3D imaging methods including multi-view stereo and DfD. Highly reflective objects pose difficulties for active 3D imaging techniques including time-of-flight and structured light. Nearly transparent objects are major challenges for both passive and active 3D imaging methods. The human face, with its complex shape and weak texture, is an even greater challenge for all existing 3D imaging methods.

The images shown in Fig. 3 are all captured under the illumination of an unpolarized light-emitting diode with its wavelength centered at 800 nm (an example of 3D reconstruction of a glass window using natural sunlight for illumination is also shown in Supplementary Section 6). In all tested scenarios, the surface normal reconstructions of the target objects using polarization information alone deviate far from the ground truth as a result of the error and ambiguity in the zenith and azimuth angle calculation. In contrast, through the multi-stage fusion of both diffraction- and polarization-based depth cues, we can reconstruct detailed and accurate 3D surfaces for all types of target objects with a normalized depth error ($\Delta z_{obj} / z_{obj}$)[18] less than 0.20% across the scenarios in Fig. 3a-c. A quantitative depth error is absent in the human face case since capturing a detailed and accurate ground truth of such a complex 3D surface using existing 3D sensors is extremely challenging.

While the prototype camera requires active near-infrared floodlighting to ensure the image brightness in the indoor scene, the EM3D framework can be transferred to white-light imaging, thus eliminating the need for active illumination, since the dispersion of DOE can be optimized and well-calibrated[34]. The absolute depth map with up to one million pixels can be generated from raw measurements within 15 seconds per scene on a laptop equipped with an Intel i7-10875H CPU and 16 GB RAM. The reconstruction can be further accelerated through operator optimization or GPU acceleration[37].

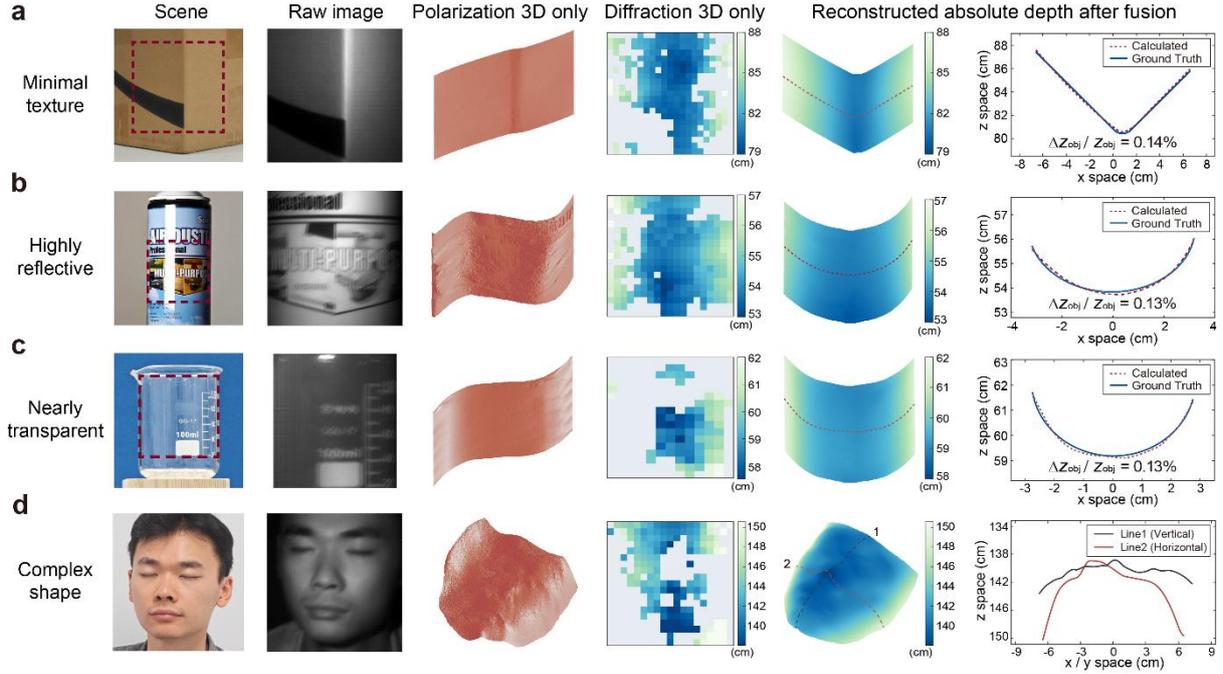

**Figure 3 | 3D imaging in extended challenging scenes. a-d,** 3D imaging results of a cardboard box with minimal texture (**a**), a highly reflective metal jar (**b**), a nearly transparent glass beaker (**c**), and a human face (the face of the 1$^{st}$-author) with a complex shape (**d**). Throughout panels **a-d**, the first column is the RGB image of the target scene, with the reconstructed region boxed by a red dashed line in panels **a-c**. The second column is the raw image captured by the monocular camera. The third column is the 3D surface reconstruction results from only the polarization-based depth cue. The fourth column is the absolute depth map calculated from only the diffraction-based depth cue, with vacancy corresponding to confidence below the threshold. The fifth column is the detailed absolute depth map combined from polarization- and diffraction-based depth cues. The sixth column shows the comparison between the reconstructed depth from the monocular camera and the ground truth with the corresponding section marked by red dashed lines in the fifth column (panels **a-c**) and vertical and horizontal contours of the reconstructed 3D surface of the human face with the corresponding section marked by grey and red dashed lines in the fifth column (panel **d**).

**Material identification by combining depth and polarization information**

Other than 3D imaging, we discover an additional, highly unique advantage of the proposed monocular camera system to perform material identification based on the surface index $n_s$ calculated from the multi-stage fusion process. An example application scenario is schematically illustrated in Fig. 4a, in which a humanoid robot aims to fetch one of three boxes placed closely on the table. The three boxes are made of iron, plastic, and ceramic, respectively, and appear too similar to be distinguished using an RGB camera, an infrared camera, or a polarization camera, as shown in Fig. 4b,c. In contrast, the monocular camera proposed here can not only provide a dense and accurate 3D point cloud of the target scene for accurate spatial localization (Fig. 4d,e); more importantly, the three boxes made of different materials are clearly distinguished by their surface indices, as depicted in Fig. 4f. We verify that the surface indices derived from the multi-stage fusion algorithm can serve as a robust indicator of material properties, which can remain consistent across various camera shooting angles and distances (Supplementary Section 7). Such a multi-modal imaging capability in a monocular camera system may greatly expand robotic vision with tight space constraints. We further show that the proposed system can be extended to other applications such as face anti-spoofing. Despite that both the depth and polarization distribution of the living human face and the fake rubber face mask are rather similar, the proposed monocular camera can clearly distinguish between a living human face from a rubber face mask based on the surface index information (Supplementary Section 8).

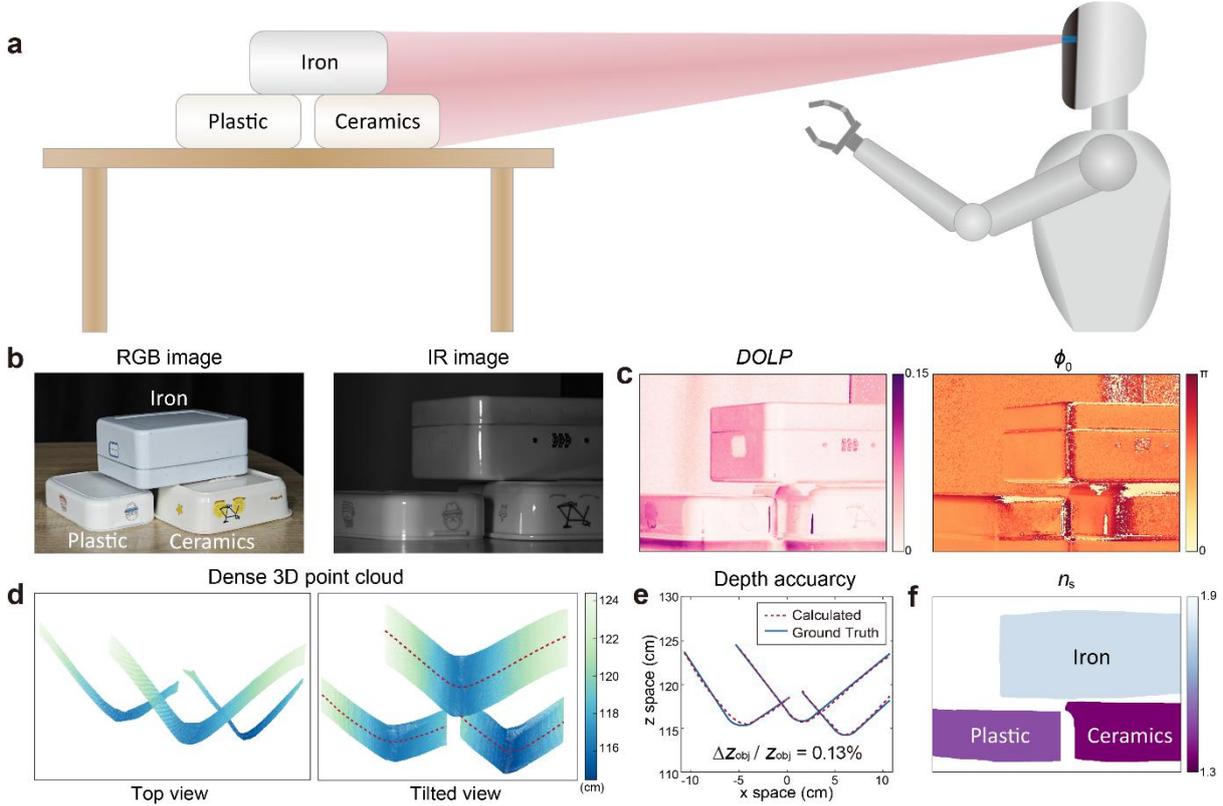

**Figure 4 | Material identification by combining depth and polarization information. a**, Schematic of an application scenario that a humanoid robot aims to fetch one of three boxes, made of different materials, placed closely on the table. **b,** RGB (left) and near-infrared (right) image of the target scene, respectively. **c,** *DOLP* (left) and polarization angle $\phi_0$ (right) analysis of the target scene, respectively, calculated from images taken by a polarization camera. **d,** Dense 3D point cloud generated by the monocular camera. **e,** The comparison of the measured depth map to the ground truth. The corresponding section is marked by red dashed lines in panel **d**. **f,** Surface indices $n_s$ of the object reconstructed from the image taken by the monocular camera.

**Summary and outlook**
To summarize, we have leveraged the vectorial wave nature of light, combining diffraction- and polarization-based depth cues, to realize a monocular camera system for high-quality snapshot 3D imaging across an extended range of scenes that have posed significant difficulties for traditional 3D imaging techniques. Moreover, we show that the synergy between depth and polarization information can unlock unique opportunities in material identification, which may greatly expand machine intelligence.

The camera employs a readily mass-producible DOE coupled with a single refractive lens for imaging. The image quality can be further improved through multi-piece diffractive-refractive hybrid lens optimization[38,39]. The form factor of the camera can be substantially shrunk via standard lens module assembly process[39] or by using more advanced wafer-level packaging[40]. We also expect that the use of polarization-sensitive metalens[23,41-45] can facilitate further system miniaturization and function expansion while eliminating the need for a polarization image sensor.

The image reconstruction algorithm developed here is fully physically driven, which is helpful for the direct evaluation and interpretation of the system performance. Looking forward, we anticipate that the complement of deep neural network or end-to-end design may further improve the system's accuracy and robustness across various scenarios under different illumination conditions[25,46]. The use of deep learning could also significantly accelerate image reconstruction, enabling real-time video-rate output. An ultra-compact, high-quality multi-dimensional imaging solution holds the potential to vastly extend the application scope of machine vision across various domains, including robotics, autonomous driving, precision metrology, and biomedical imaging[47-49].

**Methods**

**8-stage DOE fabrication**. The DOE sample is fabricated via a commercial service provided by Chengdu Zhilan Micro-nano Technology Co., Ltd. The process flow is depicted schematically in Supplementary Fig. S1. The fabrication starts with a quartz substrate with a thickness of 500 μm. To achieve an 8-stage phase distribution, we have developed a 3-step photolithography process, where each exposure step is designed to induce a height change corresponding to phase shifts of $\pi/2$, $\pi/4$, and $\pi/8$, respectively. In each exposure step, UV lithography initially inscribes the pattern for the respective layer onto the photoresist. Following this, the pattern is etched onto the quartz layer through a dry etching process. The etch depth is precisely controlled to match the phase change value of the corresponding layer. At the end of each exposure step, the photoresist is removed (details in Supplementary Section 1).

**PSF and diffraction efficiency measurement.** To measure the PSF of the diffractive-refractive hybrid lens, we construct an experimental setup as shown in Supplementary Fig. S5. The illumination source is an LED with a central wavelength of 800 nm and a bandwidth of 60 nm. The LED is focused by a convex lens with a focal length of 50 mm to a pinhole with a diameter of 30 μm, which forms a point source for depth-dependent PSF measurement, as shown in Supplementary Fig. S5. When measuring the PSF of the assembled camera, the distance between the point light source and the entrance pupil is varied, while the image distance of the camera is kept fixed.

To estimate the diffraction efficiency of the diffractive-refractive hybrid lens, a collimated laser beam with a central wavelength of 800 nm and a spot size matching the aperture size of the DOE is incident onto the DOE cascaded with a refractive lens with a focal length of 35 mm. First, to measure the power of the diffracted light $P_f$, a pinhole of 200-μm-diameter is placed in front of an optical power meter (Thorlabs PM122D), as shown in Supplementary Fig. S6a. The position of the power meter is spatially scanned and maximized near the designed focal point of the lens to determine $P_f$. Subsequently, the reference light power $P_{ref}$ is measured using the same method, with the DOE, refractive lens, and pinhole taken away, as shown in Supplementary Fig. S6b. The diffraction efficiency of the diffractive-refractive hybrid lens is 87.63%, which is calculated as $\eta = P_f / P_{ref}$.

**Data availability**
All relevant data are available in the main text, in the Supporting Information, or from the authors.

**Code availability**
The source codes demonstrating 3D reconstruction using the EM3D framework are available from https://github.com/THUMetaOptics/EM3D.

**Acknowledgment**
This work was supported by the National Natural Science Foundation of China (62135008), the National Key Research and Development Program of China (2023YFB2805800), the Beijing Municipal Science & Technology Commission, the Administrative Commission of Zhongguancun Science Park (Z231100006023006), and the China Postdoctoral Science Foundation (2023M741911).

**Author contributions**

Y.Y. and F.Z. conceived this work; Z.S. designed and characterized the DOE with assistance from Y.N; Z.S and F.Z developed the reconstruction algorithm and conducted the experiments; Z.S., F.Z. and Y.Y. wrote the manuscript. Y.Y. supervised the project.

**Competing interests**
Y.Y., F.Z. and Z.C. have submitted patent applications on technologies related to the device developed in this work. The other author declares no competing interests.

# Supplementary Information:

# Extended monocular 3D imaging

Zicheng Shen[1,#], Feng Zhao[1#,*], Yibo Ni[1], Yuanmu Yang[1]*

[1]State Key Laboratory of Precision Measurement Technology and Instruments, Department of Precision Instrument, Tsinghua University, Beijing 100084, China
# These authors contributed equally.
* Email: fengzhao@mail.tsinghua.edu.cn; ymyang@tsinghua.edu.cn

## 1. Diffractive optical element (DOE) fabrication

The fabrication process of the 8-stage DOE sample is detailed in **methods** and shown in **Fig. S1a**. The photograph of the 8-stage DOE sample used in our camera prototype is shown in **Fig. S1b**. The surface height map of the central area of the 8-stage DOE sample measured by a white light interferometer is shown in **Fig. S1c**.

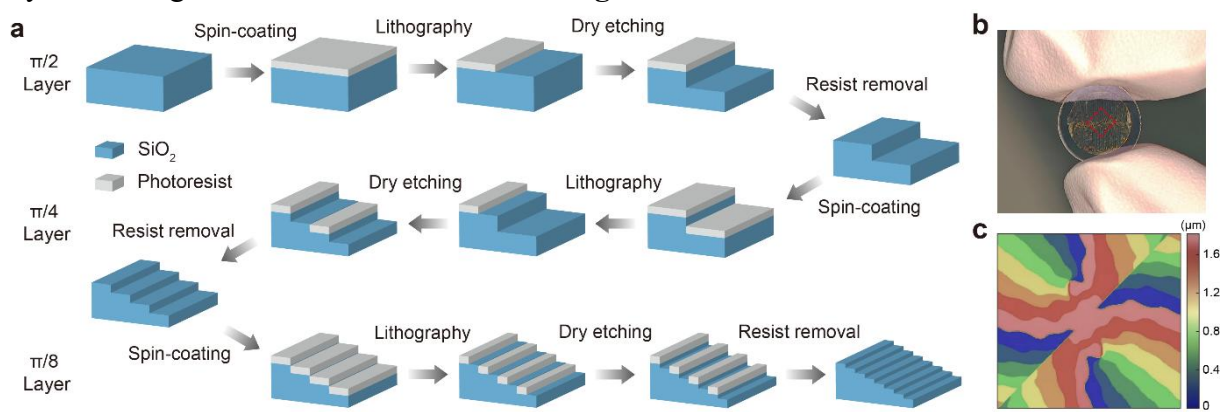

**Figure S1** | **a.** Fabrication process of the 8-stage DOE sample. **b.** Photograph of the 8-stage DOE sample fabricated by three-step photo-lithography. **c**. Measured surface height map of the central area of the 8-stage DOE sample. The corresponding area is boxed by a red dashed line in panel **b**.

In the proposed monocular 3D imaging system, the refractive lens, narrowband filter, and the polarization image sensor are all commercially available. The 8-stage DOE can also be mass-produced via nanoimprinting. The nanoimprinting process is schematically illustrated in **Fig. S2a**. We first fabricate a mold via a three-step photo-lithography process similar to that in **Fig. S1a**, but on a silicon wafer, instead of on an $SiO_2$ substrate. A 6-inch silicon wafer can accommodate hundreds of DOEs, as shown in **Fig. S2b**. After transferring the pattern on the silicon mold to a working stamp, nanoimprinting can be performed on an $SiO_2$ substrate with imprint adhesive. The photograph of the nanoimprinted 6-inch DOE wafer is shown in **Fig. S2c**, which can be subsequently diced and assembled to a hybrid diffractive-refractive lens module.

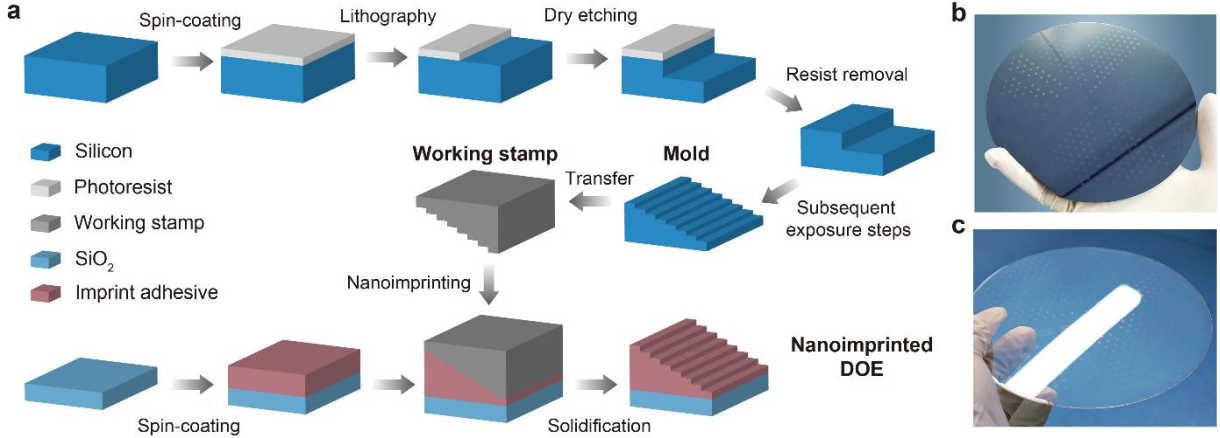

**Figure S2** | **a.** Process flow of DOE nanoimprinting. **b.** Photograph of the 6-inch silicon nanoimprinting mold. **c.** Photograph of the 6-inch nanoimprinted DOE wafer.

## 2. Double-helix point spread function (PSF) design and optimization

Conventionally, a rotating PSF is generated by a diffractive optical element (DOE) with a phase profile that is the superposition of Gauss-Laguerre modes[1-3]. Here, we employ an alternative generalized Fresnel zone approach[4] that arranges phase term with ring areas carrying spiral phase profiles of increasing topological quantum numbers toward outer rings inserted in the aperture plane of the imaging system. Compared with Gauss-Laguerre mode-based approach, the generalized Fresnel zone approach can generate a more compact rotating PSF with tailorable main lobe distance and depth of field[5]. For the first step, to generate a single rotating foci, according to the Fresnel zone approach[5,6], the phase term $\psi(u, \varphi_u)$ is given by,

$$\psi(u, \varphi_u) = \left\{ l\varphi_u \mid \left(\frac{l-1}{L}\right)^\varepsilon \leq u \leq \left(\frac{l}{L}\right)^\varepsilon, l = 1, ..., L \right\}, \quad (S1)$$

where $u$ is the normalized radial coordinate and $\varphi_u$ is the azimuth angle in the aperture plane. $[L, \varepsilon]$ are adjustable design parameters. When $L \gg 1$ and $\varepsilon = 0.5$, with the addition of a focusing lens, we can calculate the complex amplitude of the PSF based on the Fresnel integral as[4],

$$U(r, \varphi; \zeta) \approx 2\sqrt{\pi} \exp[-i\zeta/(2L)] \frac{\sin[\zeta/(2L)]}{\zeta} \times \sum_{l=1}^{L} i^l \exp[-il(\varphi - \zeta/L)] J_l\left(\frac{2\pi\sqrt{l/L}\,r}{r_0}\right), \quad (S2)$$

where $r$ is the normalized radial coordinate and $\varphi$ **is** the azimuthal angle with respect to the geometric image point of the focusing lens, which refers to the centre of the PSF when the Fresnel zone phase term is not added. $r_0$ is the radius of the in-focus diffraction spot. $\zeta$ is the defocus parameter given by,

$$\zeta = \frac{\pi}{\lambda}\left(\frac{1}{z_{\text{obj}}} - \frac{1}{z_{\text{f}}}\right) R^2, \quad (S3)$$

where $R$ is the radius of the entrance pupil of the imaging system, $z_{\text{obj}}$ is the depth of the target, and $z_{\text{f}}$ is the depth of the in-focus object plane.

According to Eq. S2, when $\zeta \ll 2\pi L$, the complex amplitude of PSF is related to $\zeta$ approximately only via the term $\exp[-il(\varphi - \zeta/L)]$. Therefore, both the complex amplitude and the intensity of the PSF, $PSF(r, \varphi; \zeta) = |U(r, \varphi; \zeta)|^2$, remain almost invariant, and rotate at the speed of $1/L$ rad per unit of $\zeta$. Consequently, the object depth is related to the rotation angle of the PSF via the defocus parameter $\zeta$. For the parameters $[L, \varepsilon]$, $L$ can be adjusted to control the rotation speed, main lobe distance, and depth of field, while $\varepsilon$ can be tuned to balance the trade-off between the main-lobe concentration and the shape-invariance during the rotation of the PSF.

In the next step, according to the generalized Fresnel zone approach[5], to generate a rotating PSF with $N$ foci, the phase term $\psi(u, \varphi_u)$ is given by,

$$\psi(u, \varphi_u) = \left\{[(l-1)N + 1]\varphi_u \Big| \left(\frac{l-1}{L}\right)^\varepsilon \leq u \leq \left(\frac{l}{L}\right)^\varepsilon, l = 1, ..., L\right\}. \tag{S4}$$

Here we choose $[N, L, \varepsilon] = [2, 12, 0.8]$ as our initial phase design to generate double-helix PSFs with suitable main lobe distance and depth of field. This phase design is shown in **Fig. S3a**. In an imaging system, such phase term can be carried by a DOE. With an additional focusing lens, double-helix PSFs are realized. For an imaging system containing a DOE with the initial phase design and a focusing lens, the numerically calculated PSFs as a function of the axis depth of a point light source are shown in **Fig. S3b**. Here the far-field distributions are calculated using the angular spectrum method[6].

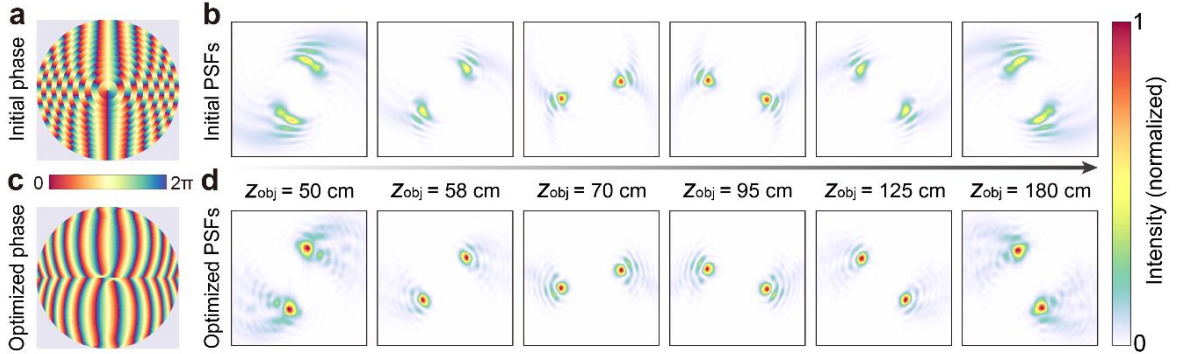

**Figure S3 | a,** Initial **generalized Fresnel zone** phase design of the DOE. **b,** Numerically calculated PSFs as a function of the axis depth of a point light source. **c,** Optimized phase design of the DOE. **d,** Numerically calculated PSFs as a function of the axis depth of a point light source.

To improve the main-lobe concentration and the shape-invariance of the double-helix PSFs generated by the DOE, thus improving the accuracy of the depth estimation as well as the quality of polarization images, we use an iterative optimization algorithm[7], with the initial generalized Fresnel zone phase term design as the input, to maximize the energy in the main-lobe of the rotating PSFs. The iterative optimization algorithm is schematically shown in **Fig. S4**.

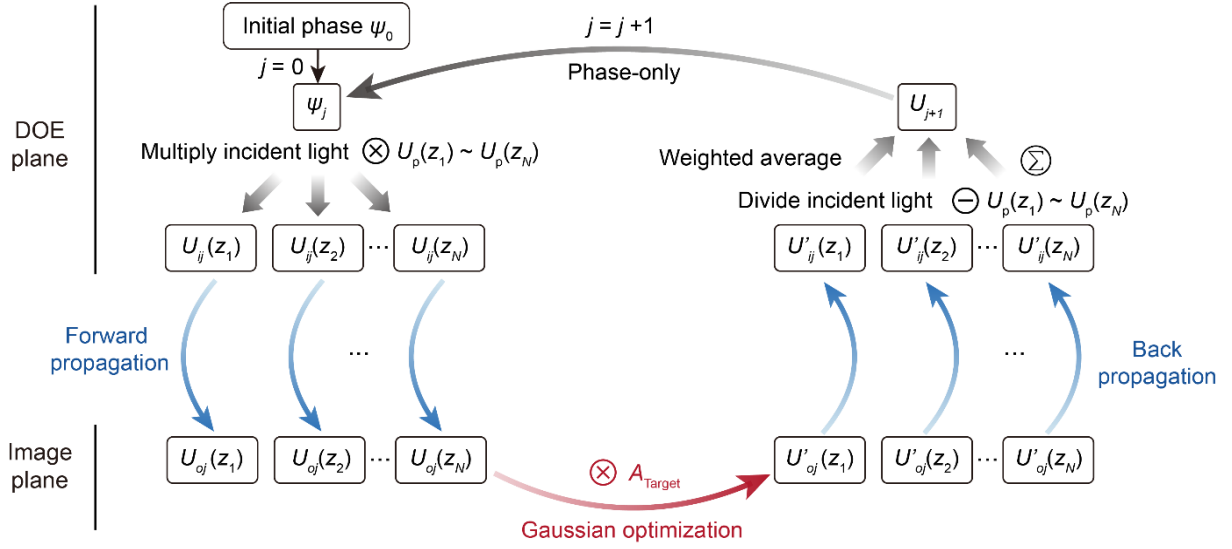

**Figure S4** | A flow chart of the iterative Fourier transform algorithm for optimizing the double-helix PSFs, where $U = A\exp(i\psi)$ represents the complex amplitude of the light field on a certain plane. The forward and inverse propagation processes are both calculated using the angular spectrum method. $U_p(z_1) \sim U_p(z_N)$ are the complex amplitude of point light sources at different axial depths ($z_1 \sim z_N$) propagated to the DOE plane.

In this iterative process, we optimize main lobe concentration and the shape-invariance of the PSFs within the 180° rotation range, which corresponds to the effective depth measurement range. Starting from the initially designed phase profile of the DOE $\psi_0$, we calculate the complex amplitude of PSFs corresponding to 9 different on-axis point light sources ($N = 9$) with depths between 50 cm and 180 cm, corresponding to $U_{oj}(z_1)$ to $U_{oj}(z_9)$ in **Fig. S4**. $A_{\text{target}}$ is a two-dimensional Gaussian function centred at the peak of the main lobe of the PSF with cut-off boundaries at 5% of the peak intensity, which is used to iteratively increase the proportion of light energy confined in the main lobe of the double-helix PSF. The optimized PSFs $U'_{oj}(z_1)$ to $U'_{oj}(z_9)$ are subsequently inversely propagated to the DOE plane and are weight-averaged after dividing the complex amplitude of the incident light. The resultant complex amplitude is given by,

$$U_{j+1} = \sum_{n=1}^{N} w_n \frac{U_{ij}(z_n)}{U_p(z_n)} \tag{S5}$$

The weights $w_n$ satisfy $\sum_{n=1}^{N} w_n = 1$. We fine-tune and set $w_1$ and $w_9$ to be relatively larger to obtain a more uniform PSF throughout the depth measurement range since PSFs corresponding to $z_1$ and $z_9$ have relatively low qualities in the initial design. The PSF amplitude is subsequently set to unity while the phase is retained for the next iteration.

This optimization process is carried out with ten iterations. The optimized DOE phase and corresponding PSFs are shown in **Fig. S3c,d**. The peak intensity and the contrast between the main lobe and the side lobe of double-helix PSFs are improved by 31% and 58% on average, respectively, within the 180° rotation range.

## 3. PSF characterization and diffraction efficiency measurement

The experimental setup for the measurement of PSFs of the assembled camera is schematically shown in **Fig. S5**, with the characterization method detailed in **methods**. The experimental setup for the measurement of the diffraction efficiency of the fabricated DOE is schematically shown in **Fig. S6**, with the characterization method also detailed in **methods**.

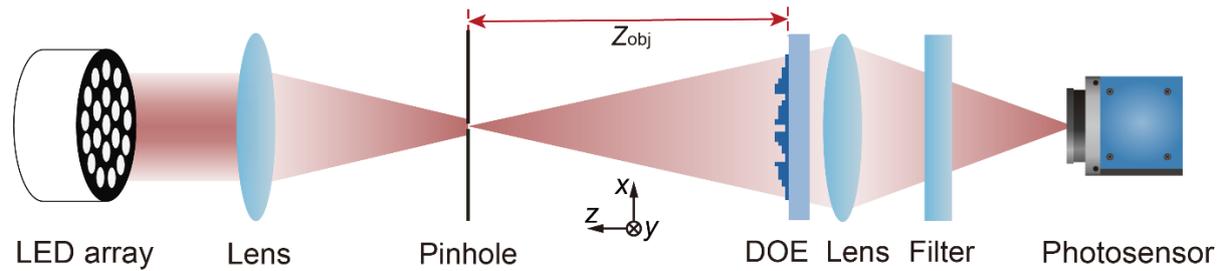

**Figure S5** | Schematic of the experimental setup for the measurement of PSFs of the assembled camera.

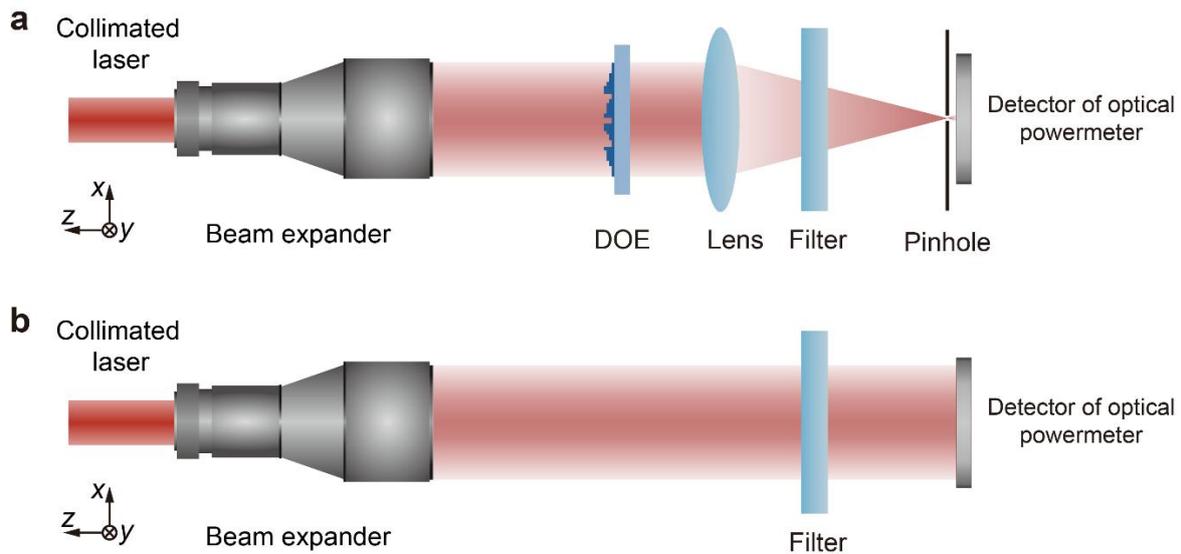

**Figure S6** | **a.** Schematic of the experimental setup for the measurement of the power of the diffracted light $P_f$. **b.** Schematic of the experimental setup for the measurement of the power of the reference light $P_{ref}$.

## 4. Working principle of shape-from-polarization (SfP)

It has been widely shown in the literature that the surface normal of an object is directly related to the polarization characteristics of light reflected from the surface[8,9]. Here we discuss in detail the method to establish the relationship between the polarization characteristics and the surface normal.

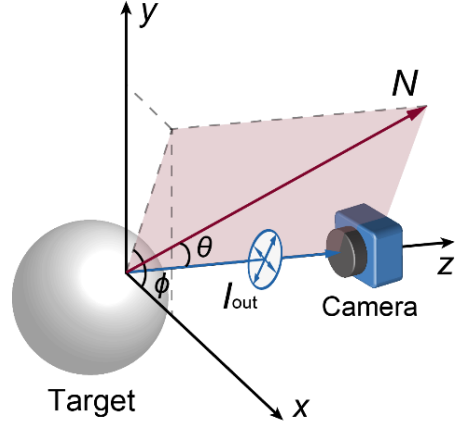

**Figure S7** | Schematic of surface normal determined from polarization information. $N$ is the surface normal of the object, $\theta$ is the zenith angle, and $\phi$ is the azimuth angle. $I_{out}$ is the reflected light received by the camera. $x$, $y$, and $z$ are spatial coordinates where the $z$-axis is along the direction of $I_{out}$. $x$-$y$ plane is parallel to the image plane.

As schematically shown in **Fig. S7**, the surface normal ($N$) of the target can be determined by the zenith angle ($\theta$) and the azimuth angle ($\phi$). The zenith angle $\theta$ represents the angle between the reflected light and the surface normal of the target object. With an unpolarized incidence, light can exit from the target object via either specular or diffuse reflection. According to the model proposed by Wolff et al[8], as shown in **Fig. S8**, the specular reflected light is directly reflected from the target surface. In contrast, the diffuse light penetrates into the material's surface layer, undergoes multiple refractions, and then transmits back into air.

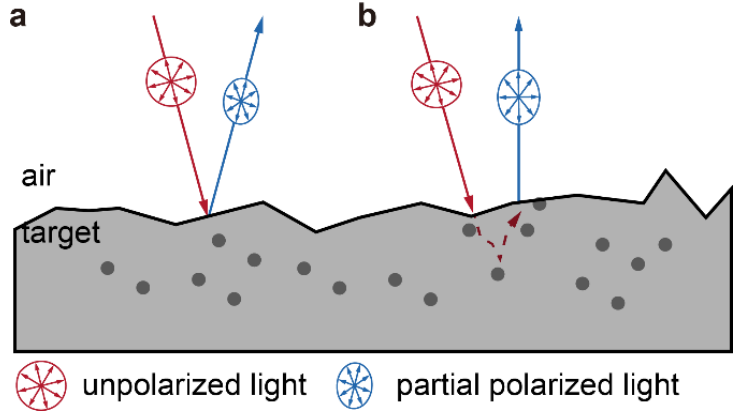

**Figure S8** |. **a**. Schematic of specular reflection from the target object. **b.** Schematic of diffuse reflection from the target object.

The degree-of-linear-polarization (*DOLP*) of light is defined as,
$$DOLP = \frac{I_{max} - I_{min}}{I_{max} + I_{min}}, \quad (S6)$$
where $I_{max}$ and $I_{min}$ are the maximum and minimum light intensity, respectively, with a linear polarizer rotated by a full cycle.

According to the Fresnel equation, the degree-of-linear-polarization of light directly reflected from the surface (specular reflection) $DOLP_r$ can be calculated as,

$$DOLP_r = \left|\frac{R_p - R_s}{R_p + R_s}\right|, \tag{S7}$$

where $R_p$ and $R_s$ refer to the reflectance of p-polarized and s-polarized light, respectively. Similarly, the degree-of-linear-polarization of light transmitted from within the surface (diffuse reflection) $DOLP_t$ can be calculated as,

$$DOLP_t = \left|\frac{T_p - T_s}{T_p + T_s}\right|, \tag{S8}$$

where $T_p$ and $T_s$ refer to the reflectance of p-polarized and s-polarized light, respectively.

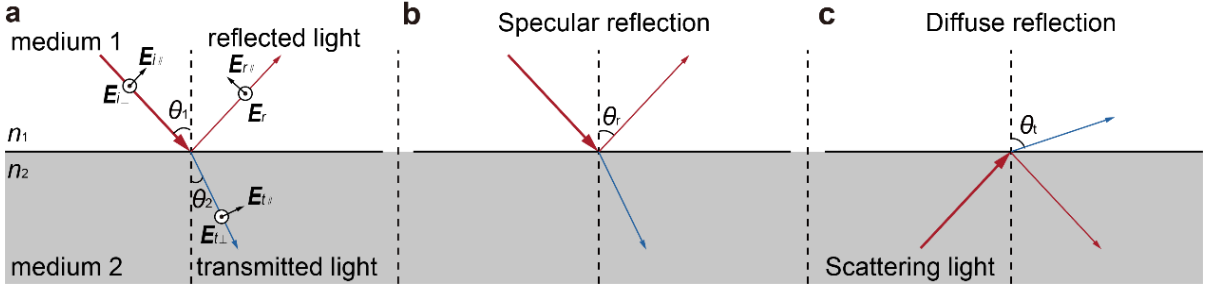

**Figure S9 | a.** The reflection and refraction processes occur at the interface between two media. $n_1$ is the refractive index of medium 1 and $n_2$ is the refractive index of medium 2. $\theta_1$ is the angle of incidence and $\theta_2$ is the angle of refraction. **b.** Schematic of the specular reflection process. $\theta_r$ is the angle of reflection and is equal to the angle of incidence and corresponds to $\theta_1$ in panel **a**. **c.** Schematic of the diffuse reflection process. $\theta_t$ is the angle of refraction and corresponds to $\theta_2$ in panel **a**.

For the geometric configuration shown in **Fig. S9a**, the Fresnel reflection and transmission coefficients can be calculated as,

$$r_s = \frac{n_1\cos\theta_1 - n_2\cos\theta_2}{n_1\cos\theta_1 + n_2\cos\theta_2}, \tag{S9}$$

$$r_p = \frac{n_2\cos\theta_1 - n_1\cos\theta_2}{n_1\cos\theta_2 + n_2\cos\theta_1}, \tag{S10}$$

$$t_s = \frac{2n_1\cos\theta_1}{n_1\cos\theta_1 + n_2\cos\theta_2}, \tag{S11}$$

$$t_p = \frac{2n_1\cos\theta_1}{n_1\cos\theta_2 + n_2\cos\theta_1}. \tag{S12}$$

The intensities of reflected and transmitted light can be subsequently calculated as,

$$R_s = r_s^2,\ R_p = r_p^2,\ T_s = t_s^2,\ T_p = t_p^2. \tag{S13}$$

As shown in **Fig. S9b,c,** the zenith angle $\theta$ of specular reflection corresponds to $\theta_r$ and the zenith angle $\theta$ of diffuse reflection corresponds to $\theta_t$. When medium 1 is air, its refractive index ($n_1$) is equal to 1. By combining Eq. S7-S13 and Snell's Law ($n_1\sin\theta_1 = n_2\sin\theta_2$), the relationship between $DOLP_r$, $DOLP_t$, and the zenith angle $\theta$ can be expressed as,

$$DOLP_r = \frac{\sqrt{\sin^4\theta\cos^2\theta\,(n^2 - \sin^2\theta)}}{[\sin^4\theta + \cos^2\theta\,(n^2 - \sin^2\theta)]/2}, \tag{S14}$$

$$DOLP_t = \frac{\left(n - \frac{1}{n}\right)^2 \sin^2\theta}{2 + 2n^2 - \left(n + \frac{1}{n}\right)^2 \sin^2\theta + 4\cos\theta\sqrt{n^2 - \sin^2\theta}}, \quad (S15)$$

where $n = n_2$ represents the refractive index of the reflecting surface. Therefore, the zenith angle $\theta$ of the surface normal can be derived from the measured $DOLP$ of reflected light and a known refractive index $n$ according to Eq. S14-S15. Nonetheless, $n$ is oftentimes an unknown parameter, and is assumed to be 1.5 in the initial step, which may lead to error in zenith angle $\theta$ determination. This issue can be addressed with the multi-stage fusion algorithm as detailed in Supplementary Section 5.

On the other hand, the azimuth angle $\phi$ ($\phi \in [0, 2\pi]$) represents the angle between the projection of the normal vector onto the plane parallel to the image plane (x-y plane in **Fig. S7**) and the horizontal direction (x-axis in **Fig. S7**). When light is reflected from the surface, the maximum intensity after passing through the polarizer corresponds to the polarization angle $\phi_0$. To find this position, one can rotate the polarizer and identify the angle at which the intensity is maximized, which corresponds to the azimuth angle $\phi$.

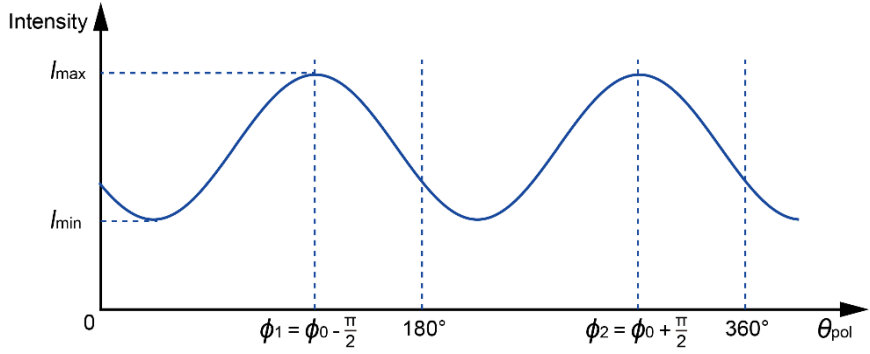

**Figure S10** | Transmitted radiance sinusoid. For diffuse reflection, the azimuth angle $\phi$ corresponds to $\theta_{pol}$ where $I_{max}$ is observed. $\phi_1$ and $\phi_2 = \phi_1 + \pi$ results in two possible surface azimuths.

As schematically shown in **Fig. S10**, if a polarizer placed in front of a camera is rotated, the measured pixel brightness can be expressed as,

$$I(\theta_{pol}) = \frac{I_{max} + I_{min}}{2} + \frac{I_{max} - I_{min}}{2} \cos(2\theta_{pol} - 2\phi_0). \quad (S16)$$

Here, $I_{max}$ and $I_{min}$ represent the maximum and minimum light intensity, respectively, with a linear polarizer rotated by a full cycle. $\phi_0$ ($\phi_0 \in [0, \pi]$) is the polarization angle of the light reflected from the surface. $\theta_{pol}$ is the rotation angle of the polarizer. By collecting multiple polarization images through the rotation of the polarizer or by using a polarization image sensor, $I_{max}$, $I_{min}$, and $\phi$ can be calculated. However, as shown in **Fig. S10**, $\phi$ and $\phi + \pi$ have the same intensity value in the polarization images, leading to the inherent ambiguity of $\pi$ radians when calculating $\phi$. This issue can also be address with the multi-stage fusion algorithm as detailed in Supplementary Section 5.

With a measured $\theta$ and $\phi$, the normal vector $N$ can be calculated as,

$$N = \begin{bmatrix} n_x \\ n_y \\ n_z \end{bmatrix} = \begin{bmatrix} \cos\theta \cos\phi \\ \cos\theta \sin\phi \\ \sin\theta \end{bmatrix} = \begin{bmatrix} \tan\theta \cos\phi \\ \tan\theta \sin\phi \\ 1 \end{bmatrix}. \tag{S17}$$

Subsequently, the three-dimensional morphology of the target object can be reconstructed by the integration of normal vectors at various points on the object's surface.

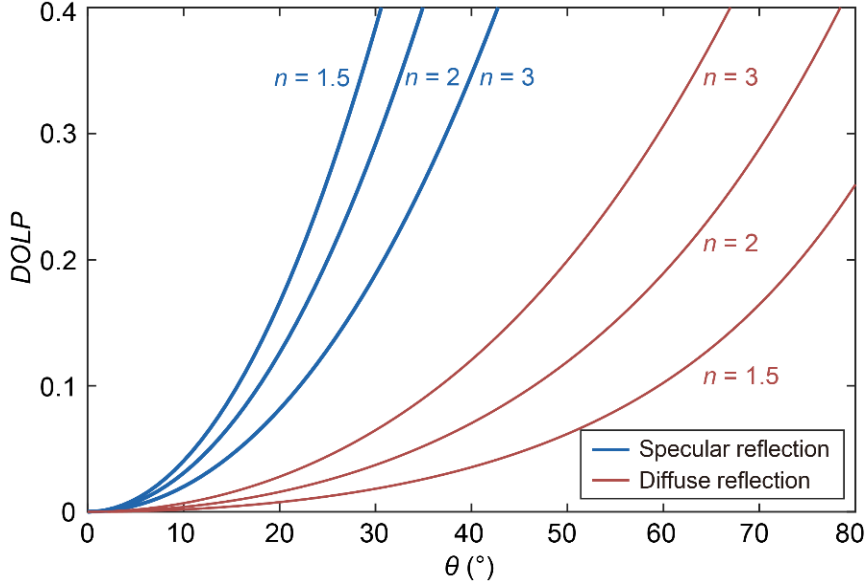

**Figure S11 | *DOLP*** as a function of $\theta$ for specular (red solid line) and diffuse (blue solid line) reflection, respectively, for different material refractive index.

Shape from specular polarization utilizes the strong polarization characteristics of the reflected light from smooth surfaces, making it primarily effective on materials like metals. However, specular reflection is highly sensitive to the direction of the light source[10]. In contrast, diffuse polarization does not require knowledge of the light source direction, while most emitted light from natural surfaces exhibits characteristics of diffuse reflection. Moreover, we find that the relationship between *DOLP* and $\theta$ of specular reflection has a similar monotonically increasing trend with that of diffuse polarization for *DOLP* < 0.4 and $\theta < 80°$, which covers the vast majority of imaging scenarios, as shown in **Fig. S11**.

Here, to simplify the image reconstruction model, we attempt to utilize only the diffuse reflection model to fit all scenarios whose reflected light can possibly include both specular and diffuse reflection. We first define a new parameter $n_s$, the surface index of the target object, a fitting parameter that replaces the refractive index $n$ in Eq. S15. By combining absolute depth information with polarization information and optimizing the fitting parameter $n_s$ with the multi-stage fusion algorithm as detailed in Supplementary Section 5, we find a close agreement between the model and the ground truth for extended challenging scenes, as is shown throughout all experiments shown in this work.

## 5. Details of the multi-stage fusion algorithm

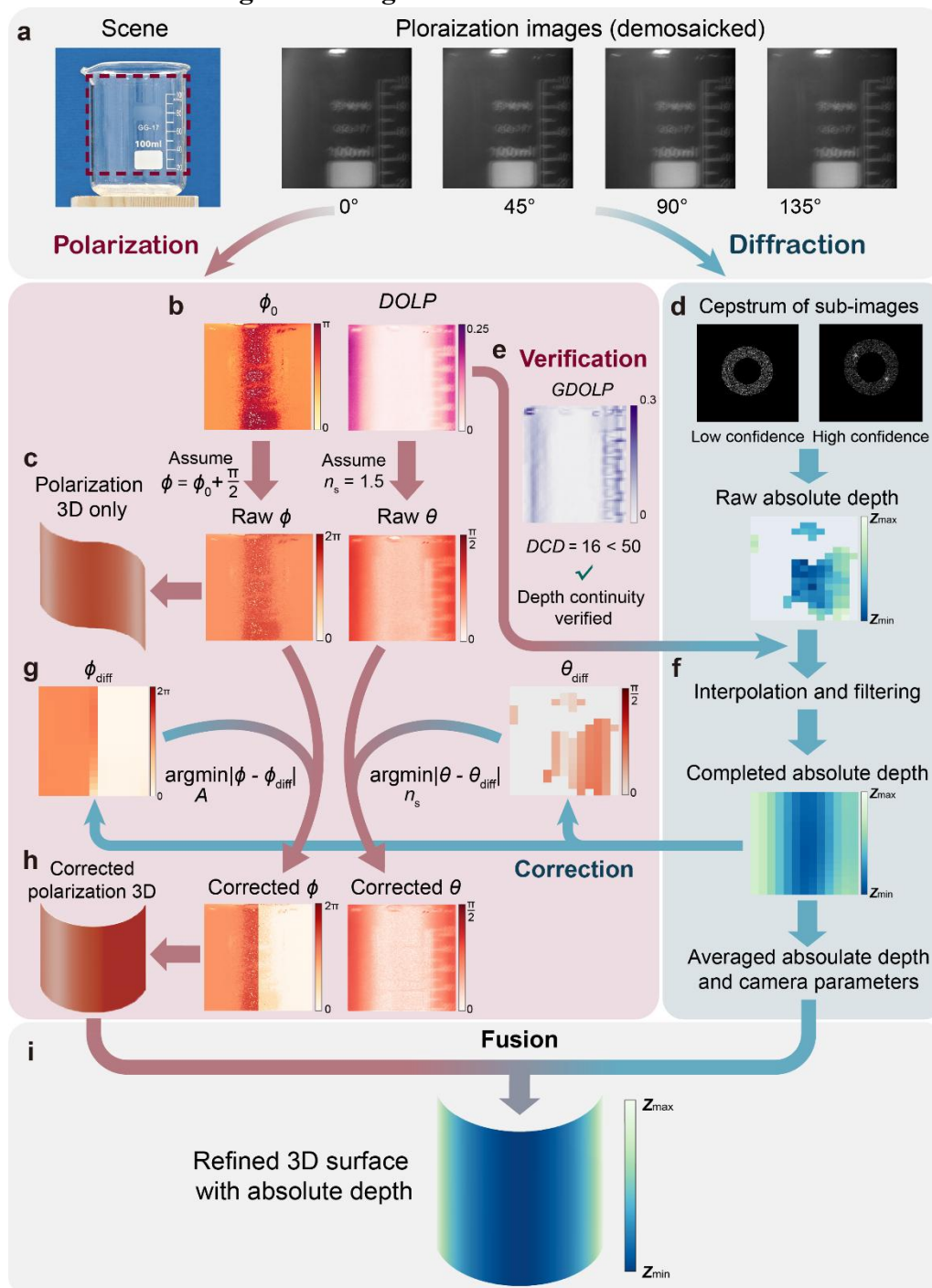

**Figure S12** | Details of the multi-stage fusion algorithm. **a**, Example scene of a nearly transparent glass beaker with the reconstructed region boxed by a red dashed line. Four polarization images are captured simultaneously by the monocular camera. **b**, Calculated polarization angle $\phi_0$ and *DOLP* map of the scene. **c**, Calculated raw azimuth angle $\phi$ and zenith angle $\theta$ from polarization information only. **d**, Calculated rough absolute depth map using diffraction-based depth cue via analyzing the power cepstrum of sub-images with a confidence threshold. **e**, Depth continuity verification by calculating *DCD*. **f**, Completed absolute depth map via interpolation and filtering. **g**, Process of resolving the ambiguity and error of raw azimuth angle $\phi$ and zenith angle $\theta$ leveraging the completed absolute depth map.

**h**, Corrected azimuth angle $\phi$ and zenith angle $\theta$. **i**, Fusion of fine 3D surface shape and the low-resolution absolute depth map to generate a fine 3D surface shape with an absolute depth of up to a million pixels

As schematically shown in **Fig. S12a**, four polarization images of the scene are first captured simultaneously using a monocular camera. Subsequently, demosaicking[11] is performed on the captured images via bilinear interpolation according to the pixel arrangement of the polarization sensor. As shown in **Fig. S12b,** using the four polarization images, we can calculate *DOLP* maps of the scene using Eq. 3 in the main text and calculate polarization angle $\phi_0$ as,

$$\phi_0 = \frac{1}{2}\arctan\left(\frac{I_0 + I_{90} - 2I_{45}}{I_{90} - I_0}\right), (\phi_0 \in [0,\pi]), \quad (S18)$$

where $I_0$, $I_{45}$, and $I_{90}$ are polarization images at 0°, 45°, and 90° polarization angles, respectively. According to Eq. 4 in the main text, a $\phi_0$ value results in two solutions of azimuth angle $\phi$ with $\pi$ difference. As shown in **Fig. S12c**, using only polarization information, we can only choose $\phi = \phi_0 \pm 90°$ blindly. On the other hand, since we have no prior information about the scene, we have to assume a surface index $n_s$ of the object and use Eq. 2 in the main text to calculate a raw zenith angle $\theta$ distribution. As a result, the surface normal reconstructed from the raw azimuth angle $\phi$ and zenith angle $\theta$ is typically incorrect.

To acquire a raw measurement of the absolute depth, we harness the diffraction-based depth cue via analyzing the power cepstrum of sub-images[12,13], as shown in **Fig. S12d**. Since absolute depth from diffraction relies on textures, too weak texture results in unreliable depth measurement, which is reflected as having no apparent peak in the power cepstrum of the sub-image. Therefore, we set a threshold on the confidence level (*CL*) to exclude the sub-images with no apparent peak in their power cepstrum. The *CL* is given by,

$$CL = \frac{P_2 + P_3}{P_1}, \quad (S19)$$

where $P_1$, $P_2$, and $P_3$ are the first, second and third largest peak values of the power cepstrum after Gaussian smoothing. The threshold of *CL* is normally set to be around 0.42. Sub-images whose *CL* are below the threshold result in void regions in the raw absolute depth map.

To fill in the void regions with continuous depth in the raw absolute depth map, we need to verify the depth trend continuity of the target object, as shown in **Fig. S12e**. By analyzing and testing various kinds of scenes, we find that the distribution of *DOLP* can be used to verify the depth continuity of images captured by the monocular camera. In detail, after down-sampling the *DOLP* image by 10 times, its second-order gradient can be calculated by applying the Laplacian operator as,

$$GDOLP(x, y) = \nabla^2(DOLP(x, y)) = \frac{\partial^2(DOLP(x, y))}{\partial x^2} + \frac{\partial^2(DOLP(x, y))}{\partial y^2}, \quad (S20)$$

where $DOLP(x, y)$ is the *DOLP* image after down-sampling and $GDOLP(x, y)$ is the second-order gradient of $DOLP(x, y)$. To evaluate the trend continuity of $DOLP(x, y)$ and thus verify the continuity of depth, we define a parameter discontinuity-of-*DOLP* (*DCD*), which is the pixel number of the largest connected area with *GDOLP* value larger than 0.15, in the *GDOLP* image. Base on the detailed camera parameters and after testing various scenarios, we set 50 as the threshold of *DCD* to decide whether the scene has continuous depth. For scenes that have continuous depth, their *DCD* values are all well below 50, as shown in **Fig. S13a**. On the other hand, for scenes that have discontinuous depth, their *DCD* values are all larger than 50, as shown in **Fig. S13b**. As experimentally demonstrated, the *DCD* threshold of 50 is small enough to avoid missing small objects (such as the tiny screw shown in the fifth row of **Fig. S13b**) with depth discontinuity, while being sufficiently large to avoid misjudging *DOLP* fluctuations within regions with continuous depth as indicators of depth discontinuity. Therefore, we can assume that scenes with *DCD* values below 50 have continuous depth, and thus fill in the void regions of corresponding raw absolute depth map. Note that when the continuous object region does not fill the full image (Fig. 3d in the main text) or multiple objects exist in the image (Fig. 4 in the main text), the continuous object region has to be segmented before performing the following steps.

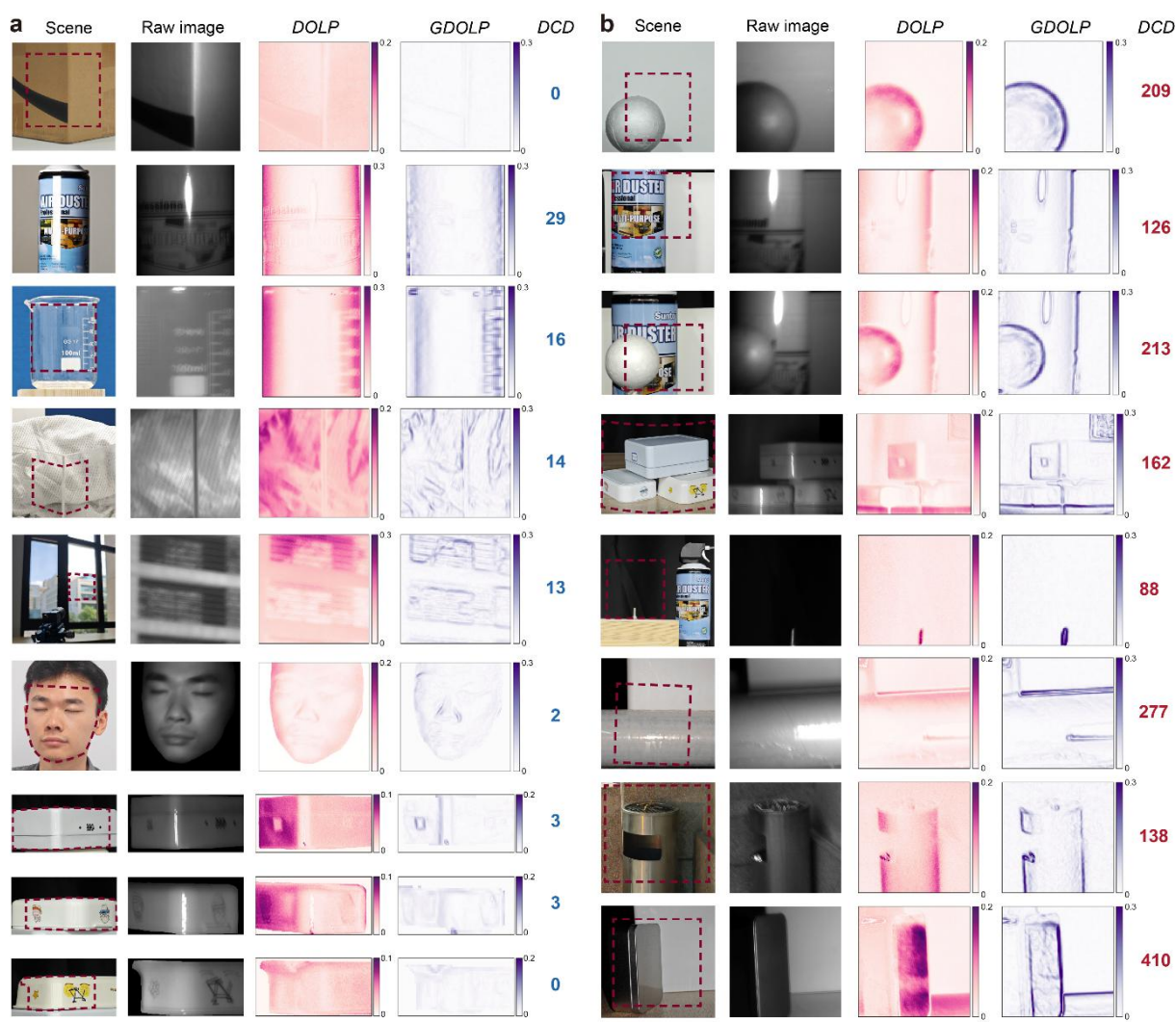

**Figure S13** | Verification of depth trend continuity for the completion of void regions in the raw absolute depth map. **a,** Scenes that have continuous depth. The first column is the RGB image of each scene with the evaluated region enclosed by a red dashed line. The second column is the raw image captured by the monocular camera. The third column is the *DOLP* of each scene. The fourth column is the *GDOLP* of each scene. The fourth column is the calculated

*DCD* of each scene. All these scenes have *DCD* smaller than 50. **b,** Scenes that have discontinuous depth. The images in the first to the fourth columns share the same meanings as those in panel **a**. All these scenes have *DCD* larger than 50.

Within the region verified to have continuous depth, we can apply interpolation and filtering to acquire a completed absolute depth map with no void region and less noisy fluctuations in depth values, as shown in **Fig. S12f**. The interpolation is performed by filling in the average of nearby non-vacant pixel values, this is followed by a Gaussian filtering to smooth out noisy fluctuations. After this step, we have a completed absolute depth map with low resolution.

To acquire a high resolution and more accurate 3D image, we need accurate azimuth angle $\phi$ and zenith angle $\theta$ distributions. The $\phi$ and $\theta$ distributions have $a \times b$ pixels each, with the corresponding matrices denoted as $\Phi$ and $\Theta$, respectively. As shown in **Fig. S12g**, we first calculate a distribution of surface normal from the absolute depth map, which results in the reference azimuth angle $\phi_{\text{diff}}$ and zenith angle $\theta_{\text{diff}}$, respectively. They don't have high resolution and pixel-wise accuracy, but yield a reliable overall direction and average value of the depth gradient. The $\theta_{\text{diff}}$ values corresponding to sub-images with *CL* below the threshold are excluded in the following correction steps. The matrix of $\phi_{\text{diff}}$ and $\theta_{\text{diff}}$ are denoted as $\Phi_{\text{diff}}$ and $\Theta_{\text{diff}}$, respectively, which are resized to have $a \times b$ pixels each, using bilinear interpolation. For the correction of azimuth angle $\phi$, we define a matrix $A$ of dimension $a \times b$, where $A \in \{0, \pi\}$. The optimization problem of correcting $\phi$ is then expressed as,

$$\widehat{A} = \underset{A}{\mathrm{argmin}} \| \Phi_{\text{diff}} - (\Phi - A) \|. \tag{S21}$$

The corrected matrix of azimuth angle $\phi$ is determined as $\Phi - A$. Therefore, by choosing the value of $A$ on each pixel to minimize the difference between $\phi$ and $\phi_{\text{diff}}$, we can correct the $\pi$ ambiguity of azimuth angle $\phi$. For the correction of zenith angle $\theta$, we need to find the correct value of surface index $n_s$. According to Eq. 2 in the main text, $\theta$ can be expressed as $\theta = f(n_s, DOLP)$, since *DOLP* of every pixel is known, this can be written as $\Theta = f(n_s)$. The optimization problem of correcting $n_s$ is then expressed as,

$$\widehat{n}_s = \underset{n_s}{\mathrm{argmin}} \| \Theta_{\text{diff}} - f(n_s) \|. \tag{S22}$$

Therefore, we can iteratively search for the optimal value of $n_s$ by minimizing the sum of absolute difference between $\theta$ and $\theta_{\text{diff}}$. Using this surface index $n_s$ jointly determined by polarization- and diffraction-based depth cues, we can calculate correct $\theta$ values according to Eq. 2 in the main text.

The corrected azimuth angle $\phi$ and zenith angle $\theta$ distribution can determine a correct distribution of surface normal of the target surface. Therefore, we can correctly reconstruct the refined 3D surface of the object via the integration method proposed by Frankot et. al[14], but with no absolute depth, as shown in **Fig. S12h**. In the final step, we harness the mean value of the completed absolute depth and the camera's intrinsic parameters to transform the values in the refined 3D surface to absolute depth values using,

$$z_a = \frac{(z_r - AVR) \times AVA \times PFOV}{PWC} + AVA, \tag{S23}$$

where $z_r$ is the relative depth value in the refined 3D surface in **Fig. S12h**, $z_a$ is the absolute depth value in the absolute depth map in **Fig. S12f**. *AVR* and *AVA* are the mean values of $z_r$ and $z_a$, respectively. *PWC* is the assumed pixel width used in the integration of surface normal. *PFOV* is the field of view angle corresponding to each pixel of the sensor, which is calibrated in experiments. Finally, we can generate a detailed 3D surface with an absolute depth of up to a million pixels by combining the refined 3D surface shape with the low-resolution absolute depth map, as shown in **Fig. S12i**.

The intermediate outputs all the experiments in the main text are shown in **Fig. S14** and **Fig. S15**.

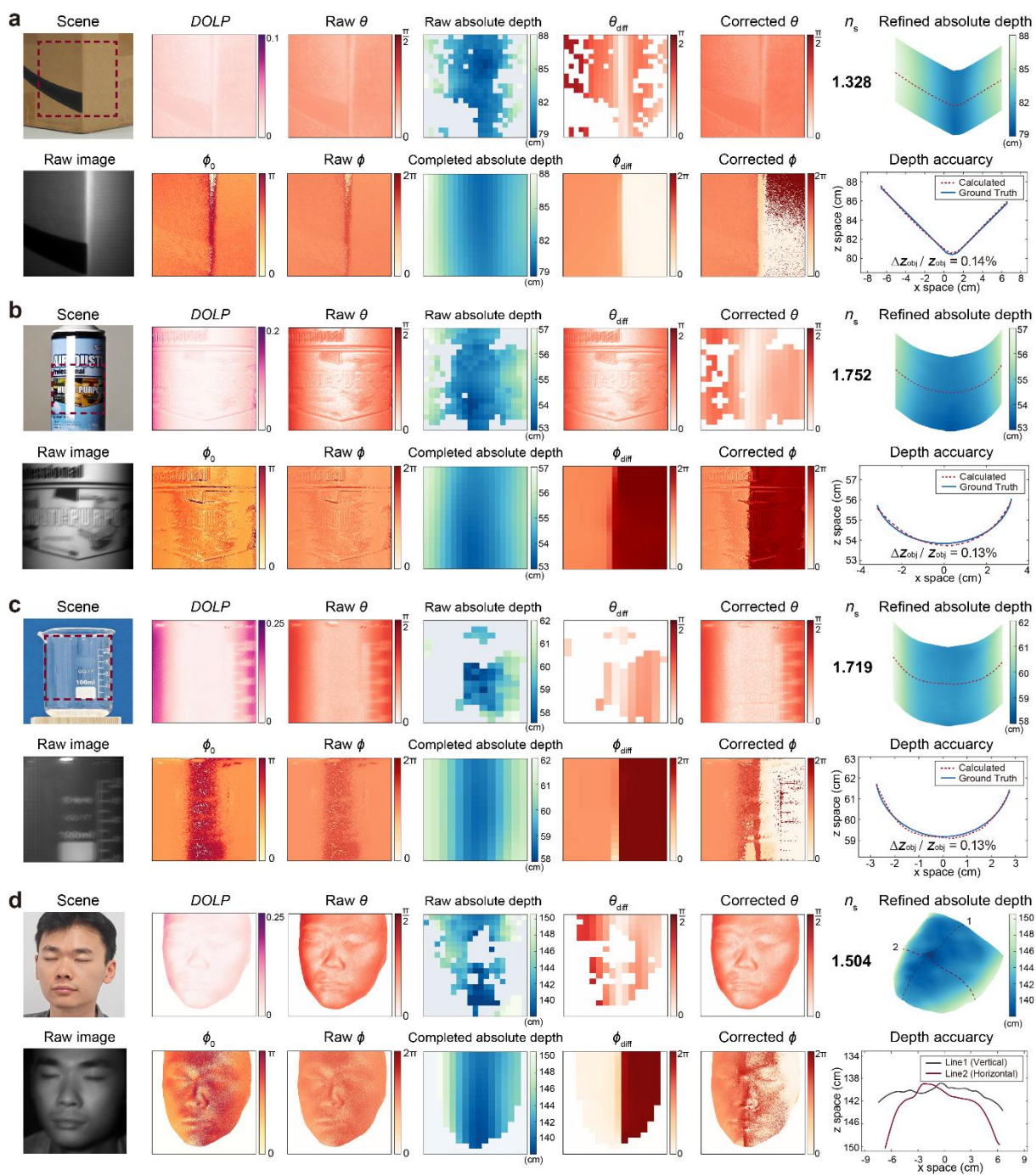

**Figure S14 | a-d,** Complete intermediate outputs of the experiments in Fig. 3 of the main

text, which include a cardboard box with minimal textures (**a**), a highly reflective metal jar (**b**), a nearly transparent glass beaker (**c**), a human face (the face of the 1st-author) with complex shape (**d**).

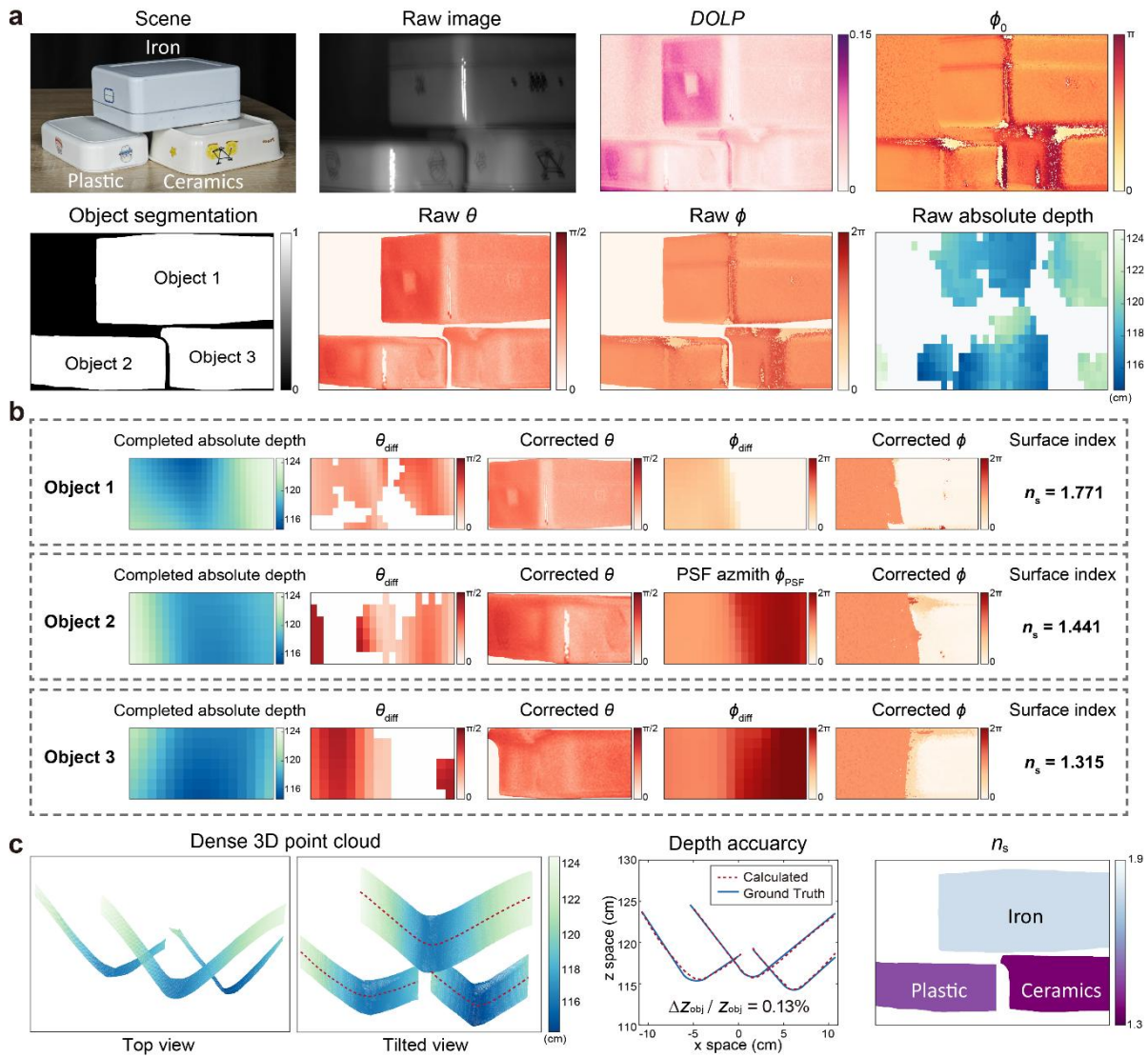

**Figure S15** | Complete intermediate outputs of the experiment in Fig. 4 of the main text. **a**, Initial analysis and segmentation of the scene. **b**, Completion and correction of 3D information from diffraction and polarization for segmented objects. **c**, 3D reconstruction and surface index results of the scene.

## 6. Additional imaging experiment on a nearly transparent scene under natural sunlight illumination

To further test the performance of our prototype camera without active illumination, we experimentally demonstrate 3D imaging of an additional nearly transparent scene under natural sunlight illumination. **Figure S16** features a nearly transparent glass window, which is under natural sunlight illumination. The glass window is accurately reconstructed with a normalized depth error ($\Delta z_{obj}/z_{obj}$) of less than 0.20%. The complete intermediate outputs of the experiment in this section are included in **Fig. S16**.

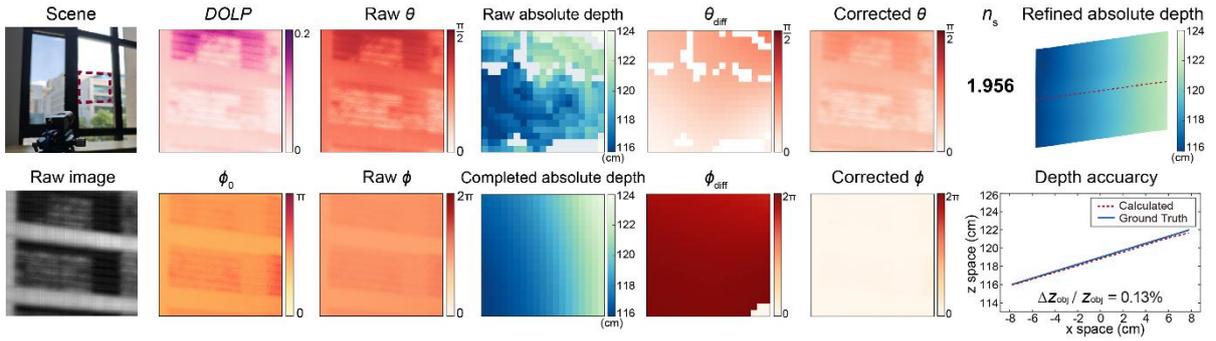

**Figure S16** | Complete algorithmic intermediate outputs of 3D imaging of a glass window under natural sunlight illumination.

### 7. Surface index dependence on camera shooting angle and distance

To ensure the effectiveness of using a calculated surface index $n_s$ for material identification, we verify the repeatability of surface index $n_s$ determination under different camera shooting angles and distances. Here we choose several different objects made of different materials as the target objects and then take images at different angles and distances using the proposed monocular camera, as is illustrated in **Fig. S17a**. Results show that shooting the same object at different angles and distances results in less than 0.01 difference of the calculated surface index $n_s$, as shown in **Fig. S17b,c**.

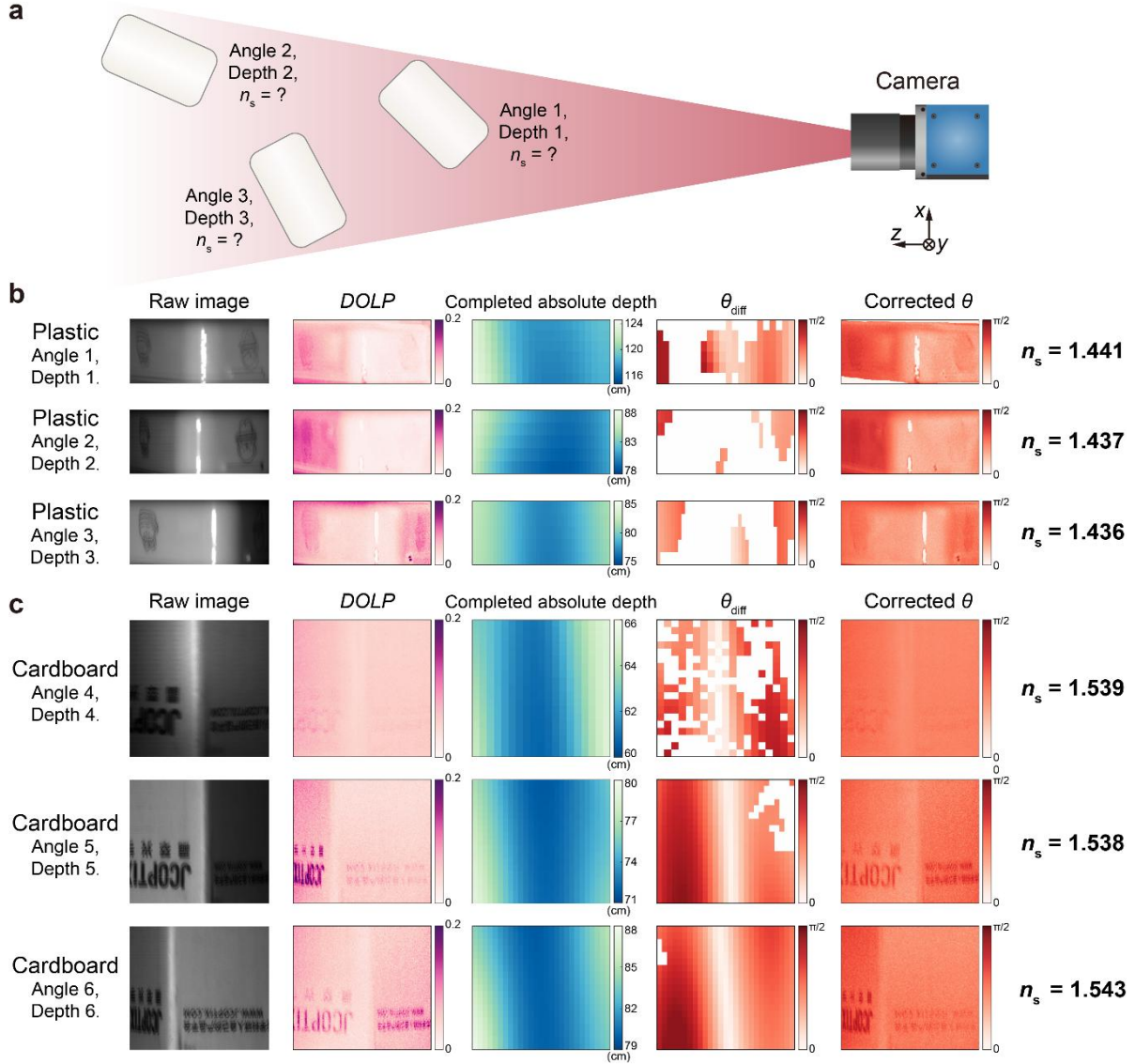

**Figure S17** | **a.** Schematic of verifying the repeatability of surface index $n_s$ calculation by shooting the same object at different angles and distances. **b,** Calculation of surface index $n_s$ for a plastic box shot at different angles and distances. The first column is the raw image captured by the monocular camera. The second column is the calculated *DOLP*. The third column is the completed absolute depth maps verified by polarization information. The fourth column is the zenith angle calculated from completed absolute depth maps. The fifth column is the zenith angle calculated from polarization information and corrected by diffraction-based zenith angle. Surface index $n_s$ of the object is calculated when correcting the polarization-based zenith angle. **c,** Calculation of surface index $n_s$ for a cardboard box shot at different angles and distances. The images in the first to the fifth columns share the same meanings as those in panel **b**.

## 8. Face anti-spoofing using the monocular camera

To explore the monocular camera's material identification capability for additional applications, we demonstrate our system can successfully identify more kinds of materials, unattainable with depth or polarization alone. As shown in **Fig. S18a**, while the two scenes, a wooden box and a cardboard box, have similar values of *DOLP*, they can be clearly distinguished by their surface indices. Furthermore, we demonstrate face anti-spoofing by showing that a living human face and a fake rubber face mask can be distinguished by their surface indices, despite their polarization and depth distributions are similar, as shown in **Fig. S18b**.

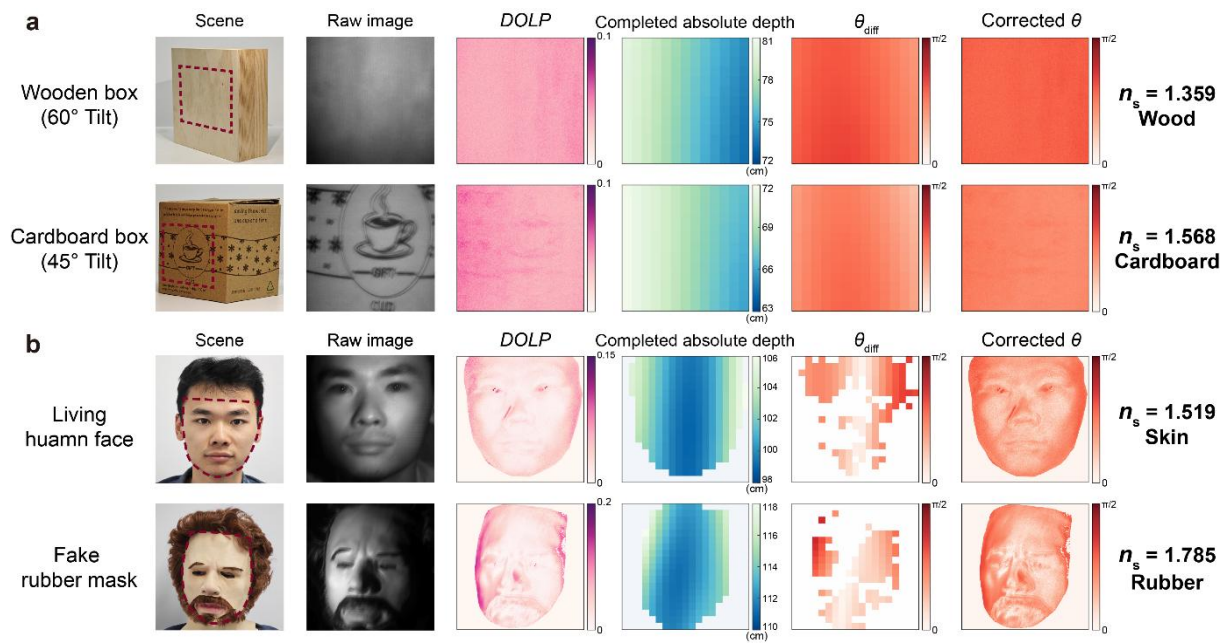

**Figure S18** | **a,** Additional experimental results on material identification. The target objects are boxes made of different materials with different tilt angles. The first column is the RGB image of each scene with the evaluated region boxed by a red dashed line. The second column is the raw image captured by the monocular camera. The third column is the calculated *DOLP* of each object, which has similar values. The fourth column is the completed absolute depth maps verified by polarization information. The fifth column is the zenith angle calculated from completed absolute depth maps. The sixth column is the zenith angle calculated from polarization information and corrected by diffraction-based zenith angle. Surface index $n_s$ of the object is calculated when correcting the polarization-based zenith angle. The material can then be distinguished using the calculated surface index $n_s$. **b,** Additional experimental results on face anti-spoofing. The images in the first to the sixth columns share the same meanings as those in panel **a**. While the *DOLP* and depth of a living human face and a fake rubber face mask have a similar distribution, the material skin and rubber can be accurately identified by the surface index $n_s$.